\documentclass[10pt,journal,compsoc,fleqn]{IEEEtran}

\ifCLASSOPTIONcompsoc

  \usepackage[nocompress]{cite}
\else

  \usepackage{cite}
\fi
\usepackage{epsfig}
\usepackage{graphicx}
\usepackage{amsmath}
\usepackage{amssymb}
\usepackage{flexisym}
\usepackage{breqn}
\usepackage{subfigure}
\usepackage{soul}
\usepackage[linesnumbered,ruled]{algorithm2e}
\usepackage{colortbl}
\definecolor{Gray}{gray}{0.85}
\usepackage{array,multirow,graphicx}
\usepackage{xfrac}
\usepackage{booktabs}
\usepackage{algorithmic}
\usepackage[flushleft]{threeparttable}
\usepackage[table]{xcolor}
\usepackage{collcell}
\usepackage{hhline}
\usepackage{pgf}
\usepackage{multirow}

\usepackage{hyperref}
\hypersetup{
    colorlinks=true,
    linkcolor=blue,
    filecolor=magenta,      
    urlcolor=cyan,
}
 
\def\colorModel{hsb} 

\newcommand\ColCell[1]{
  \pgfmathparse{#1<50?1:0}  
    \ifnum\pgfmathresult=0\relax\color{white}\fi
  \pgfmathsetmacro\compA{0}      
  \pgfmathsetmacro\compB{#1/100} 
  \pgfmathsetmacro\compC{1}      
  \edef\x{\noexpand\centering\noexpand\cellcolor[\colorModel]{\compA,\compB,\compC}}\x #1
  } 
\newcolumntype{E}{>{\collectcell\ColCell}m{0.4cm}<{\endcollectcell}}  
\newcommand*\rot{\rotatebox{90}}

\newcommand{\inner}[2]{\left\langle #1,#2 \right\rangle}

\newcommand{\real}{\ensuremath{\mathbb{R}}}

\usepackage{enumitem}

\begin{document}

\title{Sparse Coding of Shape Trajectories for Facial Expression and Action Recognition}

\author{Amor~Ben Tanfous, 
        Hassen~Drira, 
        and~Boulbaba~Ben Amor,~\IEEEmembership{Senior Member,~IEEE}.
\IEEEcompsocitemizethanks{\IEEEcompsocthanksitem A. Ben Tanfous and H. Drira are with IMT Lille Douai, CRIStAL laboratory, CNRS UMR 9189, France. B. Ben Amor is with the Inception Institute of Artificial Intelligence (IIAI), U.A.E.\protect\\
Email: omar.bentanfous@imt-lille-douai.fr}}

\IEEEtitleabstractindextext{%
\begin{abstract}

The detection and tracking of human landmarks in video streams has gained in reliability partly due to the availability of affordable RGB-D sensors. The analysis of such time-varying geometric data is playing an important role in the automatic human behavior understanding. However, suitable shape representations as well as their temporal evolution, termed trajectories, often lie to nonlinear manifolds. This puts an additional constraint (\emph{i.e.}, nonlinearity) in using conventional Machine Learning techniques. As a solution, this paper accommodates the well-known Sparse Coding and Dictionary Learning approach to study time-varying shapes on the Kendall shape spaces of 2D and 3D landmarks. We illustrate effective coding of 3D skeletal sequences for action recognition and 2D facial landmark sequences for macro- and micro-expression recognition. To overcome the inherent nonlinearity of the shape spaces, intrinsic and extrinsic solutions were explored. As main results, shape trajectories give rise to more discriminative time-series with suitable computational properties, including sparsity and vector space structure. Extensive experiments conducted on commonly-used datasets demonstrate the competitiveness of the proposed approaches with respect to state-of-the-art. 

\end{abstract}

}

\maketitle

\IEEEdisplaynontitleabstractindextext

\IEEEpeerreviewmaketitle

\IEEEraisesectionheading{\section{Introduction}\label{sec:Intro}}
\IEEEPARstart{T}{he} availability of real-time skeletal data estimation solutions~\cite{Shotton11,cao2017realtime} and reliable facial landmarks detectors~\cite{xiong2013supervised,6909636,baltrusaitis2018openface} has pushed researchers to study shapes of landmark configurations as well as their temporal evolution. For instance, 3D skeletons have been widely used to recognize human actions due to their ability in summarizing the human motion. Another example is given by the 2D facial landmarks and their tremendous use in facial expression analysis. However, human actions and facial expressions observed from visual sensors are often subject to view variations which makes their analysis  complex. Considering this non-trivial problem, an efficient way to analyze these data takes into account view-invariance properties, giving rise to shape representations often lying to nonlinear shape spaces~\cite{kendall1984shape,bryner20142d,BenAmor:2016}. David G. Kendall~\cite{kendall1984shape} defines the shape as the geometric information that remains when location, scale, and rotational effects are filtered out from an object. Accordingly, one can represent 2D landmark faces and 3D skeletons as points in the 2D and 3D Kendall's spaces, respectively. Further, when considering the dynamics of these points, the corresponding representations become trajectories in these spaces~\cite{BenAmor:2016}. However, inferencing such a representation remains challenging due to the {\it nonlinearity} of the underlying manifolds. In the literature, two alternatives have been proposed to overcome this problem for different Riemannian manifolds -- they are either \textit{Extrinsic (kernel-based)}~\cite{harandi2015extrinsic,DBLP:journals/corr/abs-1304-4344,jayasumana2013framework,6751309} or \textit{Intrinsic}~\cite{cetingul2009intrinsic,cetingul2011sparse,ho2013nonlinear,huang2016deep}. On one hand, extrinsic solutions are based on embeddings to higher dimensional Reproducing Kernel Hilbert Spaces (RKHS), which are vector spaces where Euclidean geometry applies. These methods bring the advantage that, as evidenced by kernel methods on $\real^{n}$, embedding a lower dimensional space in a higher dimensional one gives a richer representation of the data and helps capturing complex patterns. Nevertheless, to define a valid RKHS, the kernel function must be positive definite according to Mercer's theorem~\cite{973}. Several works in the literature have studied kernels on the 2D Kendall's space. For instance, the Procrustes Gaussian kernel is proposed in \cite{jayasumana2013framework} as positive definite. In contrast, to our knowledge, such a kernel has not been explored for the 3D Kendall's space. On the other hand, intrinsic solutions tend to project the manifold-valued data to a common tangent space attached to the manifold at a reference point~\cite{anirudh2015elastic,BenAmor:2016,Chellappa-CVPR-2014}. While it solves the problem of nonlinearity of the manifold of interest, this solution could introduce distortions, especially when the projected points are far from the reference point. In this work, we propose an extrinsic solution to represent 2D facial trajectories in RKHS and an intrinsic solution to model 3D actions. 
The latter brings a solution to the problem of distortions caused by tangent space approximations. In addition, we propose a comparative study of intrinsic and extrinsic solutions in the 2D and 3D Kendall's spaces.

\begin{figure*}[!ht]
  \centering
   \includegraphics[width=.8\linewidth]{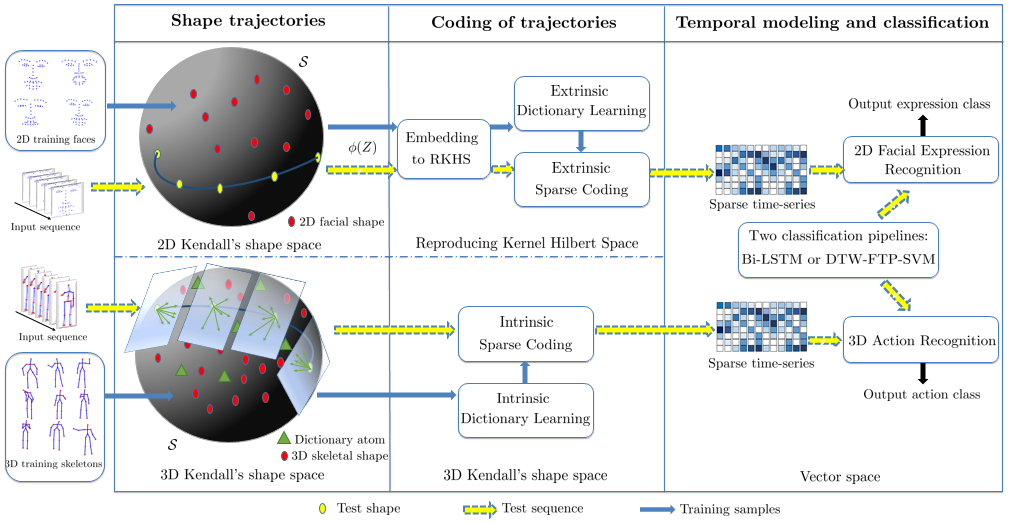}
  \caption{Overview of the proposed approaches. Sequences of 2D/3D landmark configurations are first represented as trajectories in the Kendall's shape space. A Riemannian dictionary is learned from training samples before it is used to code trajectories. This yields Euclidean sparse time-series that are temporally modeled and classified in vector space. 2D facial expression trajectories are coded using extrinsic SCDL while 3D action trajectories are coded with intrinsic SCDL.}
   \label{Fig:overview}
\end{figure*}

Motivated by the success of sparse representations in several recognition tasks \cite{7565529,harandi2015extrinsic, ho2013nonlinear}, we propose to code shape trajectories using Riemannian sparse coding and dictionary learning (SCDL). Specifically, 2D facial trajectories are coded in RKHS while sparse coding of 3D skeletal trajectories is performed in an intrinsic manner. As a main result, these coding techniques give rise to sparse times-series lying in vector spaces. In the contexts of facial expression recognition and action recognition, this brings two main advantages:~(1)~Sparse coding of shapes is performed with respect to a Riemannian dictionary. Hence, the resulting sparse times-series are expected to be more discriminative than the data themselves. In addition, they are robust to noise, knowing that SCDL is a powerful denoising tool; (2)~Using sparse time-series as discriminative features allows us to perform both temporal modeling and classification in vector space, avoiding the more difficult task of classification on the manifold. An overview of the proposed approaches is given in Figure~\ref{Fig:overview}. 
\vspace{0.1cm}

A preliminary version of this work appeared in \cite{tanfous2018coding} with an application of intrinsic SCDL in the 3D Kendall's space to model and recognize human actions. In this paper, we generalize the latter work to model and classify 2D facial expressions (micro and macro) in the 2D Kendall's space. Moreover, we will provide a comparative study between the intrinsic and extrinsic approaches in the underlying shape spaces. This will be supported by extensive experiments and discussions. In summary, the main contributions of this work are,
\begin{itemize}[leftmargin=*]
    \item A novel human action and facial expression modeling based on SCDL in Kendall shape spaces. This allows to represent shape trajectories as time-series with suitable computational properties including sparsity and vector space structure.
    \item A comparative study of intrinsic and extrinsic SCDL solutions in the 2D and 3D Kendall's spaces. To the best of our knowledge, this work is the first to apply both approaches to dynamic 2D and 3D shape related data. 
    \item Application of our framework to 3D action recognition, 2D micro- and 2D macro- facial expression recognition. Extensive experiments are conducted on seven commonly-used datasets to show the competitiveness of the proposed approach to state-of-the-art.
\end{itemize}

\vspace{0.2cm}
The rest of the paper is organized as follows. In section~\ref{sec:2}, we briefly review existing solutions of SCDL in nonlinear manifolds, geometric approaches in 3D action recognition, and some existing methods for 2D macro and micro facial expression recognition. Section~\ref{sec:3} presents the geometry of the Kendall's spaces, in addition to an embedding to RKHS that will be used to define the extrinsic SCDL solution. In section~\ref{sec:4}, we present the intrinsic and extrinsic frameworks of Riemannian SCDL. In section~\ref{sec:5}, we describe the adopted temporal modeling and classification pipelines. Experimental results and discussions are reported in section~\ref{sec:6}, and section~\ref{sec:Concl} concludes the paper and draws some perspectives.

\section{Prior Work}\label{sec:2}
In this section, we first focus our review on the extension of SCDL to nonlinear Riemannian manifolds. Then, we review geometric methods of 3D action recognition and 2D facial expression recognition.

\subsection{SCDL on Riemannian manifolds} 
Sparse representations have proved to be successful in various computer vision tasks which explains the significant interest in the last decade~\cite{7565529,harandi2015extrinsic, ho2013nonlinear}. Based on a learned dictionary, each data point can be represented as a \textit{linear combination} of a few dictionary elements (atoms), so that a squared Euclidean loss is minimized. This assumes that the data points as well as the dictionary atoms are defined in vector space (to allow speaking on linear combination). However, most suitable image features often lie to nonlinear manifolds \cite{Lui:2012}. Thus, to sparsely code these data while exploiting the Riemannian geometry of manifolds, the classical problem of SCDL needs to be extended to its nonlinear counterpart. Previous works addressed this problem~\cite{7565529,6479703,harandi2015extrinsic, 7299018,ho2013nonlinear,6751309,6619442}. For instance, a straightforward solution was proposed in \cite{6479703,Yuan2010}  by embedding the manifolds of interest into Euclidean space via a fixed tangent space at a reference point. However, this solution could generate distortions since on a tangent space, only distances to the reference point are equal to true geodesic distances. To overcome this problem, Ho ~\emph{et al.}~\cite{ho2013nonlinear} proposed a general framework for SCDL in Riemannian manifolds by working on the tangent bundle. Here, each point is coded on its attached tangent space where the atoms are mapped. By doing so, only distances to the tangent point are needed. Their proposed dictionary learning method includes an iterative update of the atoms using a gradient descent approach along geodesics. This general solution essentially relies on mappings to tangent spaces using the logarithm map operator. Although it is well defined for several manifolds, analytic formulation of the logarithm map is not available or difficult to compute for others. Therefore, some studies~\cite{harandi2015extrinsic,7299018,DBLP:journals/corr/abs-1304-4344,6751309} proposed to embed the Riemannian manifold into RKHS which depends on the existence of positive definite kernels, according to Mercer's theorem~\cite{973}. For some Riemannian manifolds, such kernels are not available in the literature which disables the extension of sparse coding to Hilbert spaces. Recently, Harandi ~\emph{et al.}~~\cite{harandi2015extrinsic} proposed to map the Grassmann manifold into the space of symmetric matrices to allow the extension of sparse coding to Grassmann manifolds. They also proposed kernelized versions of the latter approach to handle the nonlinearity of the data, similarly proposed in \cite{harandi2016sparse} for Symmetric Positive Definite matrices. In \cite{7299018}, the authors generalized sparse coding to nonlinear manifolds based on positive definite kernels. Their method was applied on three different Riemannian manifolds: the Grassmann manifold, the SPD manifold, and the 2D Kendall's shape space. In particular, kernel SCDL was applied in the latter manifold for the task of shape classification. This method will be further studied in our work in the context of 2D dynamic facial expression recognition.

\subsection{2D Facial Expression Recognition }
The problem of facial expression recognition has attracted a particular attention in the last decades due to its potential in a wide spectrum of areas. The task here is to recognize the basic emotions (\emph{e.g.}, anger, disgust, surprise, etc.) from facial videos. Early works tackled this problem by extracting hand-crafted features that combine motion and appearance from image sequences such as LBP-TOP~\cite{zhao2007dynamic} and 3D SIFT~\cite{Liu_2014_CVPR,scovanner20073}. More recent approaches exploited deep neural networks such as 3D CNNs~\cite{liu2014deeply} and RNNs~\cite{ebrahimi2015recurrent}. In~\cite{7410698}, two neural network architectures were proposed for image videos (DTAN) and 2D facial landmark sequences (DTGN) which are combined (forming DTAGN) to predict final emotions. In particular, DTGN showed to be efficient by using only 2D landmark sequences, when applied seperately. Another geometric approach was proposed in~\cite{wang2013capturing} which introduced a unified probabilistic framework based on an interval temporal Bayesian network (ITBN) built from the movements of landmark points. Aware of the small variations along a facial expression, the authors in~\cite{jain2011facial} proposed a method to capture the subtle motions within facial expressions using a variant of Conditional Random Fields (CRFs) called Latent-Dynamic CRFs (LDCRFs) on geometric features. Taking another direction, a method in~\cite{taheri2011towards} was proposed to represent 2D facial sequences as parametrized trajectories on the Grassmann manifold of 2-dimensional subspaces in $\mathbb{R}^{n}$ ($n$ is the number of landmarks) which is an affine-invariant shape representation. To capture the facial deformations, they used geodesic velocities between facial shapes and finally, classification was performed by applying LDA then SVM. In another work~\cite{Kacem_2017_ICCV}, 2D facial landmark sequences were first represented as trajectories of Gram matrices in the manifold of positive semidefinite matrices of rank $2$. A similarity measure is then provided by temporally aligning trajectories while taking into account the geometry of the manifold. This measure is finally used to train a pairwise proximity function SVM. 
\vspace{0.2cm}
Although the macro facial expression recognition problem has seen considerable advances, micro-expression recognition is still a relatively challenging task~\cite{oh2018survey}. Micro-expressions are brief facial movements characterized by short duration, involuntariness and subtle intensity. In the literature, previous methods opted for extracting hand-crafted features from texture videos such as LBP-TOP and HOOF~\cite{zheng2016relaxed}. More recently, deep learning methods were proposed to tackle the problem by applying CNNs~\cite{breuer2017deep,kim2016micro} and RNNs~\cite{kim2016micro}. To our knowledge, only the method of~\cite{choi2018recognizing} is entirely based on analyzing 2D facial landmark sequences. Their work is based on computing the point-wise distances between adjacent landmark configurations along a sequence which is stacked in a matrix. The latter was seen as an input image to a CNN-LSTM-based classifier. However, their approach was only evaluated on a synthesized dataset produced from a macro-expression dataset. In our work, we will show that we achieve state-of-the-art results on a commonly-used micro-expression dataset using only 2D landmark data.

\subsection{Human Action Recognition from 3D skeletal data}

Several approaches in the literature proposed spatio-temporal models to classify 3D action sequences. Early works extracted hand-crafted descriptors from 3D skeletal data. Popular examples include Key-Pose based descriptors~\cite{xia2012view,ofli2014sequence} and dynamics-based descriptors~\cite{zanfir2013moving,chaudhry2013bio}. More recently, deep learning was applied to recognize 3D actions. Both feed-forward neural networks such as CNNs~\cite{ke2017new,kim2017interpretable,yan2018spatial} and several variants of recurrent neural networks such as LSTM~\cite{liu2016spatio,zhu2016co,zhang2017view} were proposed. The above-mentioned approaches did not make any manifold assumptions on the data representation. However, several shape representations and their dynamics often lie to nonlinear manifolds. As a consequence, many approaches exploited the Riemannian geometry of nonlinear manifolds to analyze skeletal sequences. For instance, in \cite{Chellappa-CVPR-2014}, the authors proposed to represent skeletal motion as trajectories in the Special Euclidean (Lie) group $SE(3)^n$ (respectively $SO(3)^n$). These representations are then mapped into the correspondent Lie algebra $\mathfrak{se}(3)^n$ (respectively $\mathfrak{so}(3)^n$) which is a vector space, the tangent space attached to the Lie group at the identity, where they are processed and classified. Exploiting the same representation on Lie Groups, the authors in~\cite{anirudh2015elastic} used the framework of Transported Square-Root Velocity Fields (TSRVF)~\cite{su:AOAS:2013} to encode trajectories lying on Lie groups. They extended existing coding methods such as PCA, KSVD, and Label Consistent KSVD to these Riemannian trajectories. Another approach~\cite{BenAmor:2016} proposed a different solution by extending the Kendall's shape theory to trajectories. Accordingly, translation, rotation, and global scaling are first filtered out from each skeleton to quantify the shape. Then, based on the TSRVF, they defined an elastic metric to jointly align and compare trajectories. Here, trajectories are transported to a reference tangent space attached to the Kendall's shape space at a reference point. A common major drawback of these approaches is mapping trajectories to a reference tangent space which may introduce distortions. Conscious of this limitation, the authors in~\cite{vemulapalli2016rolling} proposed a mapping of trajectories on Lie groups combining the usual logarithm map with a rolling map that guarantees a better flattening of trajectories on Lie groups. In our work, we represent skeletal sequences as trajectories in the Kendall's shape space and to overcome its nonlinearity, we propose to code them with an intrinsic formulation of SCDL that avoids distortions caused by tangent space approximations.

\section{Preliminaries}
\label{sec:3}
In the following, we review the geometry of the Kendall's space in the case of 2D planar shapes and 3D skeletal data. Then, we describe the embedding of 2D shapes to RKHS.

\subsection{Geometry of the Kendall's shape space}
\label{KendallTheory}
Let us consider a set of $n$ landmarks in $\mathbb{R}^{m}$ ($m=2,3$). To represent its shape, Kendall \cite{kendall1984shape} proposed to establish equivalences with respect to shape-preserving transformations that are translations, rotations, and global scaling. 
Let $Z \in \mathbb{R}^{n \times m}$ represent a configuration of landmarks.
To remove the translation variability, we follow \cite{dryden-mardia} and introduce the notion of Helmert sub-matrix, a $(n-1) \times n$ sub-matrix of a commonly used Helmert matrix, to perform centering of configurations. For any $Z \in \mathbb{R}^{n \times m}$, the product $HZ \in \mathbb{R}^{(n-1) \times m}$ represents the Euclidean coordinates of the centered configuration. Let ${\cal C}_0$ be the set of all such centered configurations of $n$ landmarks in $\mathbb{R}^m$, \emph{i.e.}, 
${\cal C}_0 = \{ HZ \in \mathbb{R}^{(n-1) \times m} | Z \in \mathbb{R}^{n \times m} \}$.
${\cal C}_0$ is a $m(n-1)$ dimensional vector space and can be identified with
$\mathbb{R}^{m(n-1)}$. To remove the scale variability, we define the pre-shape space to be: ${\cal C} = \{ Z \in {\cal C}_0|  \| Z\|_F =1\}$; ${\cal C}$ is a unit sphere in $\mathbb{R}^{m(n - 1)}$ and, thus, is $m(n-1)-1$ dimensional. The tangent space at any pre-shape $Z$ is given by: $T_Z({\cal C} )= \{ V \in {\cal C}_0 | \mbox{trace}(V^TZ) = 0 \}$. To remove the rotation variability, for any $Z \in {\cal C}$, we define an equivalence class:  $\bar{Z} = \{ ZO | O \in SO(m)\}$ that represents all rotations of a configuration $Z$. The set of all such equivalence classes, ${\cal S} = \{\bar{Z} | Z \in {\cal C} \} = {\cal C}/SO(m)$ is called the {\it shape space} of configurations. The tangent space at any shape $\bar{Z}$ is $T_{\bar{Z}}({\cal S}) = \{ V \in {\cal C}_0 |  \mbox{trace}(V^T Z) =0,\ \  \mbox{trace}(V^T Z U) = 0\}\ $, where $U$ is any $m \times m$ skew-symmetric matrix. The first condition makes $V$ tangent to ${\cal C}$ and the second makes $V$ perpendicular to the rotation orbit. Together, they force $V$ to be tangent to the shape space ${\cal S}$. Assuming standard Riemannian metric on ${\cal S}$, the geodesic between two points $\bar{Z_1}, \bar{Z_2} \in {\cal S}$ is defined as:
\begin{eqnarray}
\alpha(t) = { \frac{1}{ \sin(\theta)}( \sin((1-t) \theta) Z_1 + \sin(t\theta)Z_2O^*)}, 
\end{eqnarray}
where  
$\theta =\cos^{-1}(\inner{Z_1}{Z_2O^*})$, $\inner{.}{.}$ is the inner product on $\cal S$, and $O^*$ is the optimal rotation that aligns $Z_2$ with $Z_1$:
$O^* = argmin_{O \in SO(m)} \| Z_1 - Z_2O\|_F^2$. This $\theta$ is also the geodesic distance between $\bar{Z_1}$ and $\bar{Z_2}$ in the shape space ${\cal S}$, representing the optimal deformation to connect $\bar{Z_1}$ to $\bar{Z_2}$ in ${\cal S}$. For $t = 0$, $\alpha(0) = \bar{Z_1}$ and for $t = 1$ we have $\alpha(1) = \bar{Z_2}$. The mapping of a point $\bar{Z_2} \in \cal S$ to the tangent space attached at $\bar{Z_1} \in \cal S$ is done by the logarithm map operator:
\begin{eqnarray}
\log_{\bar{Z_1}}(\bar{Z_2}) =  { \frac{\theta}{\sin(\theta)}} (Z_2O^* - \cos(\theta)Z_1).
\end{eqnarray}
 The inverse operation, e.g., exponential map, applies the shooting vector to a source shape and provides the deformed (target) shape. It is defined, for any $V \in T_{\bar{Z}}({\cal S})$, by,  
\begin{equation}
 \exp_{\bar{Z}}(V) = \left[ \cos(\theta) Z + {\frac{\sin(\theta) }{\theta}} V \right].
 \end{equation}
 Note that Kendall's shape space is a complete Riemannian manifold such that the logarithm map $\log_{\bar{Z}}$ is defined for all $\bar{Z} \in \cal S$. As a consequence, the geodesic distance between two configurations $\bar{Z_1}$ and $\bar{Z_2}$ can be computed as ${ d}_{\cal S}(\bar{Z_1},\bar{Z_2})=\Vert \log_{\bar{Z_1}}(\bar{Z_2}) \Vert_{\bar{Z_1}}$, where $\Vert . \Vert_{\bar{Z_1}}$ denotes the norm induced by the Riemannian metric at $T_{\bar{Z_1}}({\cal S})$.

\vspace{0.2cm}

\noindent \textbf{The case of planar shapes} -- For $m=2$, a 2D landmark configuration can be initially represented as a $n$-dimensional complex vector whose real and imaginary parts respectively encode the $x$ and $y$ coordinates of the landmarks. In this case, the pre-shape space is defined, after removing the translation and scale effects, as: ${\cal C}= \{ z \in \mathbb{C}^{n-1} | \| z\| =1\}$; ${\cal C}$ is a complex unit sphere of dimension $2(n-1)-1$. The rotation removal consists of defining, for any $z \in \mathbb{C}^{n-1}$, an equivalence class $\bar{z} = \{ zO | O \in SO(2)\}$ that represents all rotations of a configuration $z$. The final shape space ${\cal S}$ is the set of all such equivalence classes ${\cal S} = \{\bar{z} | z \in {\cal C} \} = {\cal C}/SO(2)$. To measure the distance between two shapes $\bar{z_1}$ and $\bar{z_2}$, we define the most popular distance on the 2D Kendall's shape space, named the full Procrustes Distance \cite{kendall1984shape}, as 
\begin{eqnarray}
\label{eq:proc_dist}
d_{FP}(\bar{z_1},\bar{z_2})=(1 - |\langle z_1,z_2\rangle|^2)^{1/2},
\end{eqnarray} 
where $\langle\cdot,\cdot\rangle$ and $|.|$ denote the inner product in $\cal S$ and the absolute value of a complex number, respectively. 

\vspace{0.2cm}

\subsection{Embedding of 2D shapes into RKHS}
\label{hilbert}
A Hilbert space $\cal H$ is a high (often infinite) dimensional vector space that possesses the structure of an inner product allowing to measure angles and distances. To define an inner product in $\cal H$, we will use a kernel function $f:(\cal S \times \cal S) \rightarrow \mathbb{R}$ which makes the resulting space a RKHS. The embedding of Kendall's space to RKHS brings the main advantage of transforming the nonlinear manifold into a vector space where one can directly apply algorithms designed for linear data. In addition, it gives a richer representation of the original data in a higher-dimensional space. This is beneficial for the specific task of SCDL which essentially relies on measures of similarities, i.e., on an inner product. However, to define a valid RKHS, the kernel function must be positive definite, according to Mercer's theorem~\cite{973}. 
For the Kendall's  space of 2D shapes, the authors of~\cite{jayasumana2013framework} have proved the positive definiteness of the Procrustes Gaussian kernel $k_P:(\cal S \times \cal S) \rightarrow \mathbb{R}$ which is defined as
\begin{eqnarray}
k_P(\bar{z_1},\bar{z_2}):=exp(-{d_{FP}}^2(\bar{z_1},\bar{z_2})/2\sigma^2),
\end{eqnarray} 
where $d_{FP}$ is the full Procrustes Distance defined in Eq.(\ref{eq:proc_dist}). This kernel is positive definite for all $\sigma \in \mathbb{R}$. In the following section, it will be used to extend SCDL to RKHS. 

\section{Riemannian coding of shapes}
\label{sec:4}
Before presenting the two Riemannian SCDL solutions, i.e., intrinsic and extrinsic, we start by recalling the classic formulation of SCDL in Euclidean space. Let $\mathcal{D}=\{d_1,d_2,...,d_N\}$ be a set of vectors in $\mathbb{R}^k$ denoting a dictionary of $N$ atoms, and $x \in \mathbb{R}^k$ a query data point. The problem of sparse coding $x$ with respect to $\mathcal{D}$ can be expressed as
\begin{equation}
\label{eq:SC-linear}
l_E(x,\mathcal{D}) =\min_{w}  \Vert x - \sum_{i=1}^{N}\left[w\right]_i d_i\Vert_2^2 + \lambda f(w),
\end{equation}
where $w \in \mathbb{R}^N$ denotes the vector of codes comprised of ${\{[w]_i}\}_{i=1}^N$, $f:\mathbb{R}^N \rightarrow \mathbb{R}$ is the sparsity inducing function defined as the $\ell_1$ norm, and $\lambda$ is the sparsity regularization parameter. Eq.~\ref{eq:SC-linear} seeks to optimally approximate $x$ (by $\hat{x}$) as a linear combination of atoms, \emph{i.e.}, $\hat{x} = \sum_{i=1}^{N}\left[w\right]_i d_i$, while tacking into account a particular sparsity constraint on the codes, $f(w)= \left\| w \right\|_1$. This sparsity function has the role of forcing $x$ to be represented as only a small number of atoms. 

\vspace{.2cm}

\noindent Given a finite set of $t$ training observations $\{x_1,x_2,...,x_t\}$ in $\mathbb{R}^k$, learning a Euclidean dictionary is defined as to jointly minimize the coding cost over all choices of atoms and codes according to:
\begin{equation}
\begin{split}
\label{eq:DL-linear}
l_E(\mathcal{D}) & =\min_{\mathcal{D},w}  \sum_{i=1}^{t} \left \Vert x_i - \sum_{j=1}^{N} [w_i]_j d_j \right \Vert_2^2 + \lambda f(w_i).
\end{split}
\end{equation}
To solve this non-convex problem, a common approach alternates between the two sets of variables, $\mathcal{D}$ and $w$, such that: (1) Minimizing over $w$ while $\mathcal{D}$ is fixed is a convex problem (\emph{i.e.}, sparse coding). (2) Minimizing Eq.~\ref{eq:DL-linear} over $\mathcal{D}$ while $w$ is fixed is similarly a convex problem.

\subsection{Extrinsic approach}
The SCDL algorithms depend on the notion of inner product. In the following, we will discuss how it can be easily extended to RKHS. 

\subsubsection{Extrinsic Sparse Coding}
A closed-form solution of extrinsic sparse coding is proposed in \cite{harandi2015extrinsic}. To derive it, let us first define $\phi : \cal S \rightarrow \cal H$ a mapping to RKHS induced by the kernel $k(\bar{z_1},\bar{z_2}) = \phi (\bar{z_1})^T \phi (\bar{z_2})$, where $\bar{z_1},\bar{z_2} \in \cal{S}$ . For a query shape $\bar{z}\in \cal{S}$, extending Eq.~\ref{eq:SC-linear} to RKHS yields
\begin{equation}
\label{eq:SC-kernel}
l_{\cal H}(\bar{z},\mathcal{D}) = \min_{w}  \Vert \phi (\bar{z}) - \sum_{i=1}^{N}\left[w\right]_i \phi (\bar{d}_i)\Vert_2^2 + \lambda f(w),
\end{equation}
with $\sum_{i=1}^{N}\left[w\right]_i = 1$. In Eq.~\ref{eq:SC-kernel}, since the sparsity term depends entirely on $w$, only the reconstruction term needs to be kernelized. Expanding the latter gives
{\setlength\mathindent{3pt}
\begin{equation*}
\Vert \phi (\bar{z})-\sum_{i=1}^{N}\left[w\right]_i \phi (\bar{d}_i)\Vert_2^2 = \phi (\bar{z})^T \phi (\bar{z}) 
\end{equation*} 
\begin{equation*}
= - 2\sum_{i=1}^{N}\left[w\right]_i {\phi (\bar{d}_i)}^T \phi (\bar{z}) + \sum_{i,j=1}^{N}\left[w\right]_i \left[w\right]_j {\phi (\bar{d}_i)}^T \phi (\bar{d}_j) 
\end{equation*}
\begin{equation}
\label{eq_kSC}
= k(\bar{z},\bar{z}) - 2 w^T k(\bar{z},D) + w^T K(D,D)w,
\end{equation}} 
where $k(\bar{z},\cal D)$ is the $N$-dimensional kernel vector computed between the query $\bar{z}$ and the dictionary atoms, and $K(\cal D,\cal D)$ is the $N \times N$ kernel matrix computed between the atoms. An efficient solution of kernel sparse coding can be obtained by considering $U\Sigma U^T$ as the SVD of the symmetric positive definite kernel $K(D,D)$, and $k(\bar{z},\bar{z})$ as a constant term (independent on $w$). Thus, Eq.~\ref{eq_kSC} can be written as the least-squares problem in $\mathbb{R}^N$: $\min_{w} \Vert  \Tilde{z} - \Tilde{D}w \Vert_2^2$, where $\Tilde{D} = \Sigma^{1/2}U^T$ and $\Tilde{z} = \Sigma^{-1/2}U^{T} k(\bar{z},D)$ (we refer to \cite{harandi2015extrinsic} for the proof). In this work, this approach is applied in the Kendall's shape space by using the kernel defined in subsection~\ref{hilbert}.

\subsubsection{Extrinsic Dictionary learning}
Similarly to Euclidean dictionary learning, the extrinsic Riemannian formulation is based on an alternating optimization strategy to update weights and atoms. While the first step is obtained with extrinsic sparse coding presented above, the second is presented in what follows. Given the codes from the first step, the problem of dictionary learning can be viewed as optimizing Eq.~\ref{eq:SC-kernel} over $\cal D$. The main idea here is to represent $\cal D$ as a linear combination of the training samples $Y$ in RKHS, according to the Representer theorem~\cite{Schlkopf2001AGR}. The resulting weights for the $M$ training samples are stacked in a $M \times N$ matrix $V$, which gives $\phi (D) = \phi (Y) V$. Since only the first term in Eq.~\ref{eq:SC-kernel} depends on $\cal D$, the problem of dictionary update can be written as $U(V) = \Vert \phi (Y) - \phi (Y)VW \Vert_2^2$, where W is the $N \times M$ matrix of sparse codes obtained from the first step. The latter can be expanded to
{\setlength\mathindent{1pt}
\begin{equation*}
U(V) = Tr(\phi (Y)(I_M - VW)(I_M - VA)^T \phi (Y)^T)
\end{equation*}
\begin{equation*}
= Tr(K(Y,Y)(I_M - VW - W^T V^T + VWW^T V^T)).
\end{equation*}} 
To obtain the updated dictionary that is now defined by $V$, the gradient of $U(V)$ is zeroed out w.r.t $V$. This gives $V = (WW^T)^{-1}W = W ^{\dagger}$, where $^{\dagger}$ is the pseudo-inverse operator.

\subsection{Intrinsic approach}
\label{sec:Intrinsic}
To deal with the nonlinearity of Kendall's space, a common approach opted for projecting manifold-valued data to a tangent space at a reference point (e.g., the mean shape). However, such a projection only results in first-order approximation of the data. The latter can be distorted, especially if points are far from the tangent point. In what follows, we will show how this problem can be avoided in the intrinsic formulation of SCDL.

\subsubsection{Intrinsic Sparse Coding}

\begin{figure}[ht]
 \centering
   \includegraphics[width=.95\linewidth]{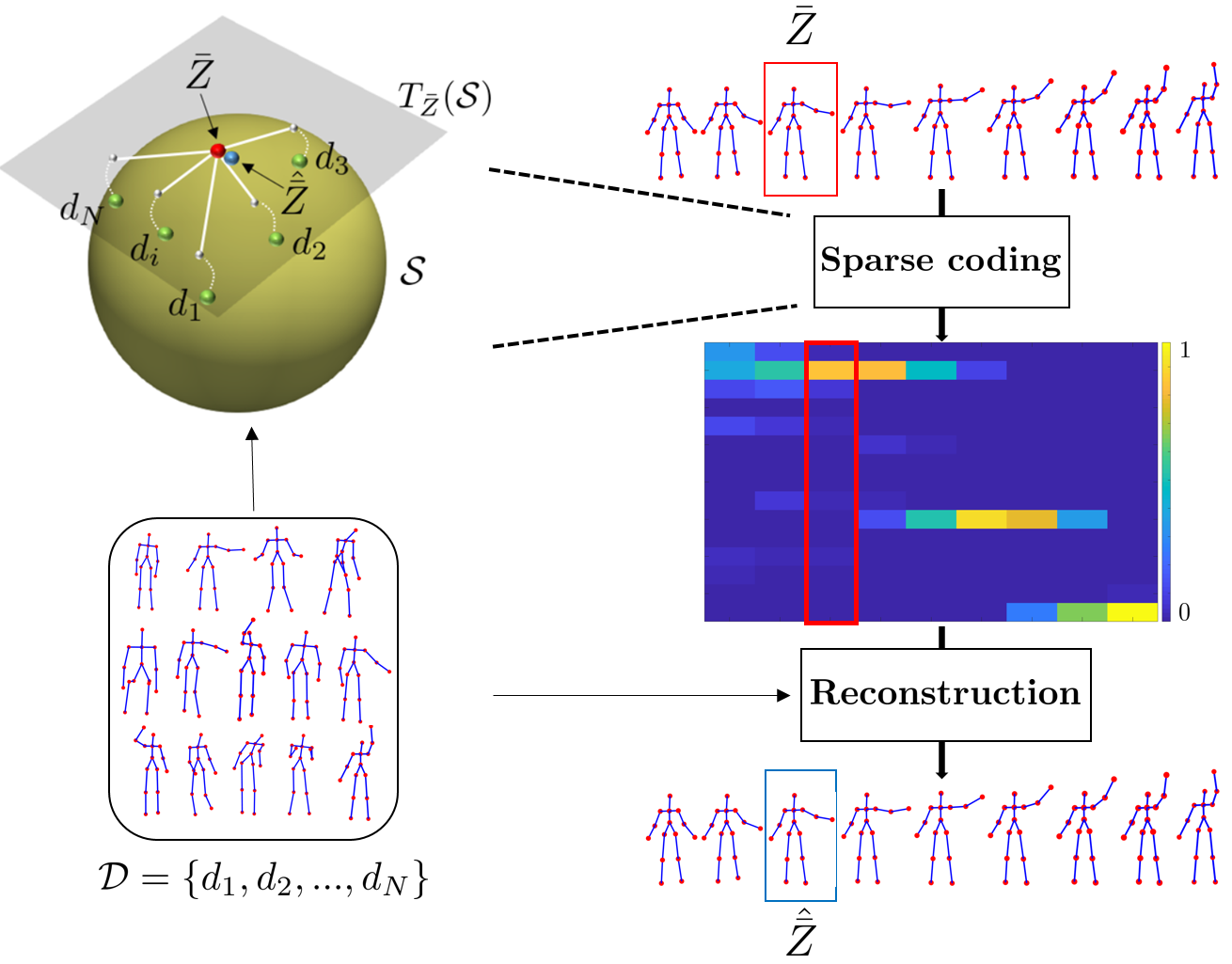}
  \caption{Illustration of intrinsic sparse coding in the Kendall's shape space. Given a dictionary $\mathcal{D}=\{d_i\}_{i=1}^N$, a skeletal trajectory is coded as a smoothly-varying sparse time-series. Using $\mathcal{D}$, the original trajectory can be reconstructed with the weighted Karcher mean algorithm.}
   \label{Fig:coding}
\end{figure}

Let $\mathcal{D}=\{\bar{d_1},\bar{d_2},...,\bar{d_N}\}$ be a dictionary on $\cal S$, and similarly the query $\bar{Z}$ is a point on $\cal S$. Accordingly, the problem of sparse coding involves the geodesic distance defined on $\cal S$ and, thus, becomes
\begin{equation}
\label{eq:SC1-manifold}
l_{\cal S}(\bar{Z},\mathcal{D}) =\min_w (d_{\mathcal{\cal S}} (\bar{Z},F(\mathcal{D},w))^2 + \lambda f(w)).   \\
\end{equation}
Here, $F:\mathcal{\cal S}^N \times \mathbb{R}^N \rightarrow \mathcal{\cal S}$ denotes an encoding function that generates the approximated point $\hat{\bar{Z}}$ on $\cal S$ by combining atoms with codes. Note that in the special case of Euclidean space, $F(\mathcal{D},w)$ would be a linear combination of atoms. However, in the Riemannian manifold $\cal S$, we have forsaken the structure of vector space which makes the linear combination of atoms lying on $\cal S$ no longer applicable, since the approximated $\hat{\bar{Z}}$ may lie out of the manifold. An interesting alternative is the intrinsic formulation of Eq.~\ref{eq:SC1-manifold}, when considering that $\cal S$ is a complete Riemannian manifold, thus, the geodesic distance $d_{\cal S}(\bar{Z},\bar{d})=\Vert \log_{\bar{Z}}(\bar{d}) \Vert_{\bar{Z}}$ (as explained in section \ref{KendallTheory}). As a consequence, the cost function in~\ref{eq:SC1-manifold} can be written as
\begin{equation}
\label{eq:SC_kendall}
l_{\cal S}(\bar{Z},\mathcal{D}) =\min_{w}  \Vert \sum_{i=1}^{N}\left[w\right]_i \log_{\bar{Z}}(\bar{d_i}) \Vert_{\bar{Z}}^2 + \lambda f(w), 
\end{equation}
where $\log_{\bar{Z}}$ denotes the logarithm map operator that maps each atom $\bar{d} \in \cal S$ to the tangent space $T_{\bar{Z}}(\cal S)$ at the point $\bar{Z}$ being coded, and $\Vert . \Vert_{\bar{Z}}$ is the norm induced by the Riemannian metric at $T_{\bar{Z}}(\cal S)$. Mathematically, this allows to partially compensate the lack of vector space structure on $\cal S$, as illustrated in Figure~\ref{Fig:coding}.
To avoid the solution $w=0$, we imposed in Eq.~\ref{eq:SC_kendall} an important additional affine constraint defined as $\sum_{i=1}^{N}\left[w\right]_i = 1$. By this formulation of sparse coding, we only compute distances to the tangent point, hence we avoid the commonly induced distortions when working in a reference tangent space. By substituting the logarithm map by its explicit formulation in Eq.~\ref{eq:SC_kendall}, we have 
\begin{equation}
\label{eq:SC_final}
l_{\cal S}(\bar{Z},\mathcal{D}) =\min_{w}  \Vert \sum_{i=1}^{N}\left[w\right]_i {\frac{\theta}{\sin(\theta)}} (d_{i}O^* - \cos(\theta)Z) \Vert_{\bar{Z}}^2 + \lambda f(w). 
\end{equation}
In practice, Eq.~\ref{eq:SC_final} is computed by first finding the optimal rotation $O^*$ between $Z$ and each atom $d_i$ via the Procrustes algorithm \cite{kendall1984shape}. Then, we solve for $w$ using the state-of-the-art CVXPY optimizer \cite{cvxpy}. 

\subsubsection{Intrinsic Dictionary Learning }
Learning a discriminative dictionary $\mathcal{D}$ typically yields accurate reconstruction of training samples and produces discriminative sparse codes. We propose a dictionary learning algorithm based on the sparse coding framework described above. 
Let $\mathcal{D}=\{\bar{d_1},\bar{d_2},...,\bar{d_N}\}$ be a dictionary on $\cal S$, and similarly $\{\bar{Z_1},\bar{Z_2},...,\bar{Z_t}\}$ is a set of $t$ training samples on $\cal S$. Similarly to the sparse coding problem, we introduce in Eq.~\ref{eq:DL-linear} the geodesic distance defined on $\cal S$ computed as $d_{\cal S}(\bar{Z},\bar{d})=\Vert \log_{\bar{Z}}(\bar{d}) \Vert_{\bar{Z}}$. As a consequence, the problem of dictionary learning on Kendall's shape space is written as
\begin{equation}
\label{eq:DL-manifold}
\min_{\mathcal{D},w} \sum_{i=1}^{t} \left\Vert  \sum_{j=1}^{N} [w_i]_j \log_{\bar{Z_i}}\bar{d_j}  \right \Vert_{\bar{Z_i}}^2 + \lambda f(w_i),
\end{equation}
with the important affine constraint $\sum_{j=1}^{N}\left[w\right]_j = 1$. Similar to the Euclidean case, the optimization problem can be solved by iteratively performing sparse coding while fixing $\mathcal{D}$, and optimizing $\mathcal{D}$ while fixing the sparse codes. 
\begin{figure}[!ht]
  \centering
   \includegraphics[width=.75\linewidth]{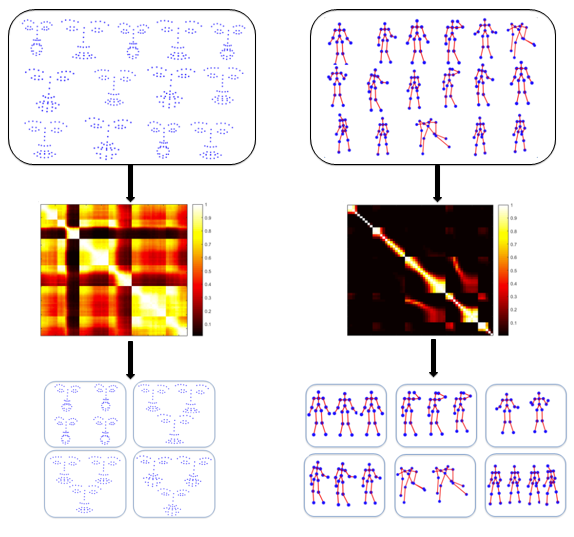}
  \caption{Illustration of the proposed clustering approach. 2D facial shapes (respectively 3D skeletons) are mapped from the 2D (respectively 3D) Kendall's space to RKHS by computing the inner product matrix from the data. Bayesian clustering is then applied on this matrix to construct the final clusters whose number is automatically inferred. }
   \label{Fig:clustering}
\end{figure}

\subsection{Kernel clustering of shapes for dictionary learning}
\label{sec:clustering}
The performance of SCDL depends on the number of the dictionary elements $N$, and an empiric choice of $N$ can be time consuming, especially when it comes to large datasets. As a solution, we propose an initialization step that enables an automatic inference on $N$ and accelerates the convergence of the dictionary learning algorithm. To this end, we propose to cluster the training shapes by adapting the Bayesian clustering of shapes of curves method proposed in \cite{Zhang2015171}. 
\noindent In Figure~\ref{Fig:clustering}, we show the main steps of the proposed clustering approach. First, an inner product matrix is computed from the training data based on the kernel function defined in subsection~\ref{hilbert}. Note that in the 3D case, this kernel is positive definite for only certain values of the kernel parameter $\sigma$. Thus, its empiric choice is required to seek positive definiteness. The inner product matrix is then modeled using a Wishart distribution. To allow for an automatic inference on the number of clusters, prior distributions are carefully assigned to the parameters of the Wishart distribution. Then, posterior is sampled using a Markov chain Monte Carlo procedure based on the Chinese restaurant process for final clustering. We refer the reader to \cite{Zhang2015171} for further details.

\vspace{0.2cm}

\noindent \textbf{Dictionary initialization} -- Given a set of training samples on $\cal S$, the idea is to select $N$ representatives to initialize the dictionary. This is done in two main steps: (1) Clustering of shapes as described above; (2) Generating atoms from each cluster such that they well describe the intra-cluster variability.  In the second step, for each cluster, we propose to perform principal geodesic analysis (PGA), first proposed by \cite{fletcher2004principal}, to obtain the best representatives of the cluster. Specifically, we map all cluster elements to the tangent space of the mean shape $T_{\bar{\mu}}(\cal S)$. Then, we perform principal component analysis (PCA) in this vector space. Finally, the resulting vectors from all the clusters are mapped to $\cal S$ to represent the initial atoms of $\cal D$. Note that an advantage of performing PGA in each cluster rather than in the whole training set is to avoid the problematic case of having points in the manifold that are far from the tangent point.

\section{Temporal modeling and classification}
\label{sec:5}
Let $\{\bar{Z_1},\bar{Z_2},...,\bar{Z_L}\}$ be a sequence of landmark configuration representing a trajectory on $\cal S$. As described in section~\ref{sec:4}, we code each skeleton $\bar{Z_i}$ into a sparse vector of  codes $w_i \in \mathbb{R}^{N}$ with respect to a dictionary $\cal D$ ($\cal D$ is given a particular structure described later on in this section). As a consequence, each trajectory is mapped to a $N$-dimensional function of sparse codes and the problem of classifying trajectories on $\cal S$ is turned to classifying $N$-dimensional sparse codes functions in Euclidean space, where any traditional operation on Euclidean time-series (\emph{e.g.}, standard machine learning techniques) could be directly applied. Several methods in the literature tend to process and classify time series \cite{anirudh2015elastic,BenAmor:2016,Chellappa-CVPR-2014,vemulapalli2016rolling}. In our work, we adopt two different classification schemes to perform action and facial expression classification: (1) A pipeline of Dynamic Time Warping (DTW), Fourier Temporal Pyramid (FTP), and one-vs-all linear SVM. Thus, we handle rate variability, temporal misalignment and noise, and classify final features, respectively; (2) Bidirectional Long short-term memory (Bi-LSTM) which is an extension of the traditional LSTM that represents each sequence backwards and forwards to two separate recurrent networks, providing context from both the future and past \cite{graves2005framewise}. 

\vspace{0.2cm}

\noindent \textbf{Dictionary structure} -- In the context of classification, one may exploit the important information of data labels to construct more discriminative feature vectors. To this end, we propose to build {\it class-specific} dictionaries, similarly to~\cite{guha2012learning}. Formally, let $S$ be a set of labeled trajectories on $\cal S$ belonging to $q$ different classes $\{c_1,c_2,...,c_q\}$, we aim to build $q$ class-specific dictionaries $\{D_1,D_2,...,D_q\}$ in $\cal S$ such that each $D_j$ is learned using skeletons belonging to training sequences from the corresponding class $c_j$. In this scenario, coding a query skeletal shape $\bar{Z}\in \cal S$ is done with respect to each $D_{j,1\leq j \leq q}$, independently. As a result, $q$ vectors of codes are obtained. These vectors are finally concatenated to form a global feature vector $W$.

\section{Experiments}
\label{sec:6}
To evaluate the proposed modeling approaches, we conducted extensive experiments on three applications: 2D macro facial expression recognition, 2D micro-expression recognition, and 3D action recognition. We further provide a comparative evaluation on the two SCDL frameworks that we used in the context of these applications. 

\vspace{0.2cm}

\noindent \textbf{Experimental Settings and Parameters} -- In all experiments, the values of the kernel parameter $\sigma$ and the sparsity regularization parameter $\lambda$ were chosen empirically. We classified the final time-series based on the two classification schemes presented in section~\ref{sec:5}. In the first scheme, i.e., DTW-FTP-SVM, we used a six-level FTP and fixed the value of SVM parameter C to 1. In the second scheme, we train the network with one Bi-LSTM layer, with the exeption of NTU-RGB+D dataset where two layers were used. The minimization is performed using Adam optimizer and the applied probability of dropout is 0.3. The value of neuron size was chosen empirically for each dataset.

\subsection{2D Facial Expression Recognition}
In this application, we extract 49 facial landmarks from human faces in 2D and with high accuracy using a state-of-the-art facial landmark detector~\cite{6909636}. We first represent the sequences of landmarks as trajectories in the Kendall's shape space. Extrinsic SCDL is then applied to produce sparse time-series that are finally classified in vector space. We evaluate this approach on two different 2D facial expression recognition problems: the macro and micro.
\subsubsection{Macro-Expression Recognition}
The task here is to recognize the basic macro emotions, \emph{e.g.}, fear, surprise, happiness, etc. To this end, we applied our approach on two commonly-used datasets namely the Cohn-Kanade Extended dataset and the Oulu-CASIA dataset. Our obtained results are then discussed with respect to state-of-the-art approaches as well as to intrinsic SCDL. For both datasets, we followed the commonly-used experimental setting in \cite{ELAIWAT2016152,7410698,6909622,6247974} consisting on a 10-fold cross validation.

\begin{itemize}
    \item \textbf{Cohn-Kanade Extended (CK+)} dataset \cite{5543262} consists of 327 image sequences performed by 118 subjects with seven emotion labels: \textit{anger}, \textit{contempt}, \textit{disgust}, \textit{fear}, \textit{happiness}, \textit{sadness}, and
\textit{surprise}. Each sequence contains the two first temporal phases of the expression, \emph{i.e.}, neutral and onset (with apex frames).

\item \textbf{Oulu-CASIA} dataset \cite{article} includes 480 image sequences performed by 80 subjects. They are labeled with one of the six basic emotions (those in CK+, except the \textit{contempt}). Each sequence begins with a neutral facial expression and ends with the expression apex.
\end{itemize} 
\vspace{0.2cm}

\begin{figure}[!ht]
  \centering
   \includegraphics[width=.98\linewidth]{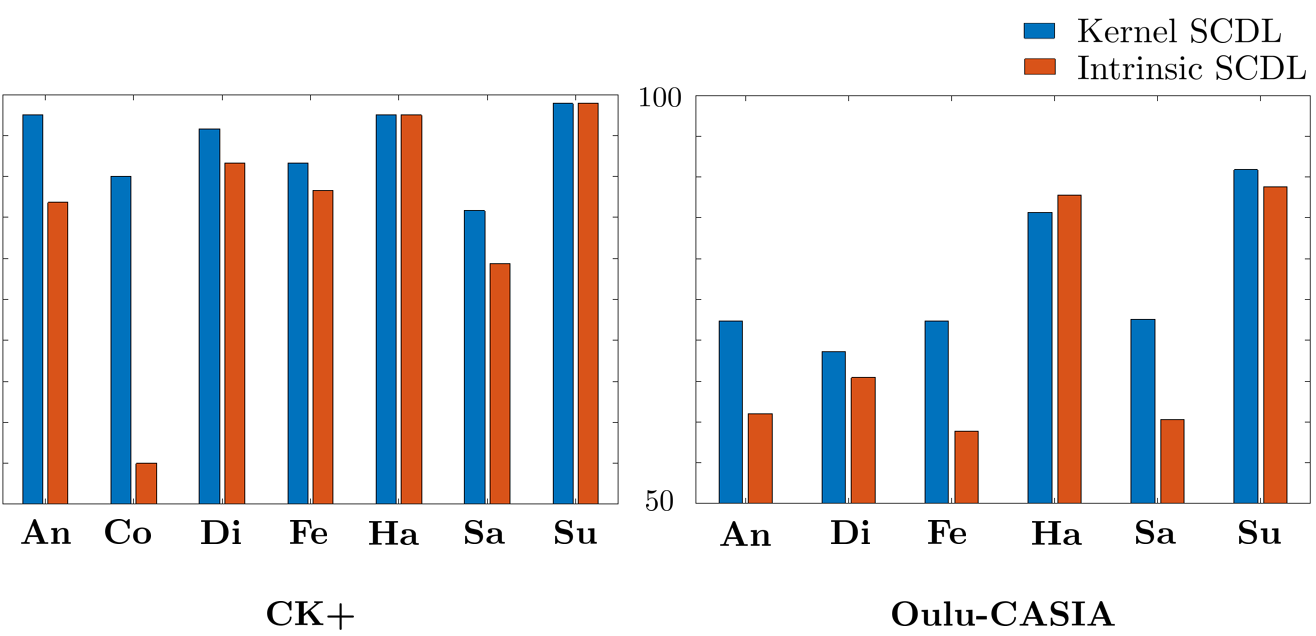}
  \caption{Recognition accuracy achieved for each emotion class in the CK+ (left) and the CASIA (rigth) datasets, and comparison between extrinsic and intrinsic SCDL approaches. }
   \label{Fig:class_acc}
\end{figure}

\noindent \textbf{Results and discussions} -- Table~\ref{tab:CK+} gives an overview of the obtained results on both datasets. Overall, our approach achieved competitive results compared to the literature. For instance, our best result on CK+ (obtained with Bi-LSTM) is by $1.52\%$ lower than the best state-of-the-art result obtained by the method of \cite{7410698}. The latter is based on two neural network architectures trained on image videos and facial landmark sequences. However, when using only the landmark architecture (DTGN), our approach obtained a higher accuracy. Similarly, on Oulu-CASIA, our best result is lower than DTAGN and higher than DTGN. On the other hand, the method of \cite{Kacem_2017_ICCV} achieved a better performance on both datasets compared to our method. Comparing the confusion matrices, the same method seems to better recognize the \textit{sadness} expression while our method is clearly more efficient in recognizing the \textit{contempt} expression. This will be further discussed later on. From Fig.~\ref{Fig:class_acc} and the confusion matrix in Table~\ref{tab:conf_casia}, we can observe that the two expressions: \textit{happiness} and \textit{surprise} are well recognized in the two datasets while the main confusions happened in the two expressions: \textit{ fear} and \textit{sadness}, conforming to state-of-the-art results~\cite{Kacem_2017_ICCV,7410698}. Besides, we highlight the superiority of extrinsic SCDL compared to intrinsic SCDL. The first is performed in RKHS which is a higher dimensional vector space. This helps capturing complex patterns in facial expressions and identifying subtle differences between similar expressions. For instance, an interesting observation could be seen for the \textit{contempt} expression. As stated in \cite{5543262}, the latter is quite subtle and it gets easily confused with other, strong emotions. For this expression, the recognition accuracy obtained with intrinsic SCDL is $55\%$, compared to $90\%$ obtained with extrinsic SCDL, as shown in Figure~\ref{Fig:class_acc}. We argue that this remarkable improvement comes from the mapping to RKHS for the same reasons mentioned above. This observation has pushed us to further evaluate the performance of our approach in the task of micro-expression recognition. 
\begin{table}[!htb]
  \centering
  \caption{Comparison with state-of-the-art on CK+ and Oulu-CASIA datasets. $^{(A)}$: Appearance-based approaches; $^{(G)}$:~Geometric approaches; $^{(R)}$:~Riemannian approaches; Last row: our approach.}\label{tab:CK+}
  \begin{footnotesize}
\begin{tabular}{|{c}||{c}||{c}|}
\hline
\textbf{Method} & \textbf{CK+} & \textbf{Oulu-CASIA}
\\
\hline 
  
  $^{(A)}$ CSPL \cite{6247974} & 89.89  & --\\
  $^{(A)}$ ST-RBM \cite{ELAIWAT2016152} & 95.66 & --\\
  $^{(A)}$ STM-ExpLet \cite{6909622} & 94.19 & 74.59\\
  $^{(G)}$ ITBN \cite{wang2013capturing} & 86.30  & --\\
  $^{(G)}$ DTGN \cite{7410698} & 92.35 & 74.17 \\
  $^{(A+G)}$ DTAGN \cite{7410698} & \textbf{97.25} & 81.46 \\
  $^{(R)}$Shape velocity on Grassmannian \cite{taheri2011towards} & 82.80 & -- \\
  $^{(R)}$Shape traj. on Grassmannian \cite{Kacem_2017_ICCV} & 94.25 & \textbf{80.0}\\
  $^{(R)}$Gram matrix trajectories \cite{Kacem_2017_ICCV} & \textbf{96.87} & \textbf{83.13} \\ 
\hline

  $^{(R)}$\textbf{Intrinsic SCDL} (SVM) & 91.26 & 70.37\\
  $^{(R)}$\textbf{Intrinsic SCDL} (Bi-LSTM) & 89.43 & 70.24 \\
\hline
  $^{(R)}$\textbf{Extrinsic SCDL} (SVM) & 95.62 & \textbf{77.06}\\
  $^{(R)}$\textbf{Extrinsic SCDL} (Bi-LSTM) & \textbf{95.73} & 73.09\\
\hline
  \end{tabular}
   \end{footnotesize}
\end{table}


\newcommand\items{6}   
\arrayrulecolor{white} 
\begin{table}[!htb]
  \centering
  \caption{Confusion matrix on the Oulu-Casia dataset.}
  \label{tab:conf_casia}
\begin{tabular}{cc*{\items}{|E}|}
\multicolumn{1}{c}{} &\multicolumn{1}{c}{} &\multicolumn{\items}{c}{\textbf{Predicted}} \\ 
\multicolumn{1}{c}{} & 
\multicolumn{1}{c}{} & 
\multicolumn{1}{c}{\rot{Angry}} & 
\multicolumn{1}{c}{\rot{Disgust}} &
\multicolumn{1}{c}{\rot{Fear}} & 
\multicolumn{1}{c}{\rot{Happy}} & 
\multicolumn{1}{c}{\rot{Sadness}} &
\multicolumn{1}{c}{\rot{Surprise}} \\
\hhline{~*\items{|-}|}
\multirow{\items}{*}{\rotatebox{90}{\textbf{Actual}}} 
& Angry & 72.33  & 12.33 & 2.11 & 1 & 12.22 & 0 \\ \hhline{~*\items{|-}|}
& Disgust & 14.22  & 68.56 & 6.22 & 3.0 & 8.0 & 0 \\ \hhline{~*\items{|-}|}
& Fear & 5.22  & 2.0 & 72.33 & 5.11 & 9.33 & 6.0 \\ \hhline{~*\items{|-}|} 
& Happy & 4.0  & 0 & 9.33 & 85.67 & 1.0 & 0 \\ \hhline{~*\items{|-}|}
& Sadness & 15.22  & 4.11 & 6.11 & 2.0 & 72.56 & 0 \\ \hhline{~*\items{|-}|}
& Surprise & 0  & 2.1 & 5.0 & 0 & 2.0 & 90.89 \\ \hhline{~*\items{|-}|}
\end{tabular}

\end{table}

\arrayrulecolor{black} 


\subsubsection{Micro-Expression Recognition}
\label{sec:micro}
Micro expressions are brief facial movements characterized by short duration, involuntariness and subtle intensity. We argue that to recognize them, in contrast to macro-expressions, we are more interested in detecting subtle shape changes along a sequence. To this end, we applied the extrinsic SCDL framework as in macro-expression recognition, and to further detect the subtle deformations, we computed displacement vectors as the difference between successive sparse codes of $L$-dimensional time-series. Then, the resulting sequences of length $L-1$ are finally used for classification. We evaluate our approach on the most commonly-used dataset, namely CASME II.

\vspace{0.2cm}

\begin{itemize}

\item  \textbf{CASME II dataset} \cite{yan2014casme} contains 246 spontaneous micro-expression video clips recorded from 26 subjects and regrouped into five classes: happiness, surprise, disgust, repression and others. We performed classification based on the commonly used Leave-one-subject-out protocol.
\end{itemize}

\noindent Recall that previous methods that tackled the problem of micro-expression recognition are appearance-based (\emph{i.e.}, using texture images) and to our knowledge, only \cite{choi2018recognizing} has studied the problem using 2D facial landmarks. However, their approach was only evaluated on a synthesized dataset produced from CK+ (macro) videos, by selecting the three first frames of an expression, then interpolating between them. For this reason, we compare our results with respect to appearance-based methods, as shown in Table~\ref{tab:casme}. 
\begin{table}[!htb]
  \centering
  \caption{Recognition accuracy on CASME II dataset and comparison with state-of-the-art methods. In the first column: $^{(A)}$:~Appearance-based approaches. $^{(R)}$:~Riemannian approaches. Last row:~our approach. }\label{tab:casme}
  \begin{footnotesize}
\begin{tabular}{|{c}|{c}|}
\hline
\textbf{Method} & \textbf{Accuracy (\%)}  \\
\hline 
  
  $^{(A)}$ STCLQP\cite{huang2016spontaneous} & 58.39 \\
  $^{(A)}$ CNN \cite{breuer2017deep} & 59.47 \\
  $^{(A)}$ CNN (LSTM) \cite{kim2016micro} & 60.98 \\
  $^{(A)}$ LBP-TOP, HOOF \cite{zheng2016relaxed} & 63.25 \\
  $^{(A)}$ Optical Strain \cite{liong2016spontaneous} & 63.41 \\
  $^{(A)}$ DiSTLBP-IIP \cite{huang2016spontaneous} & \textbf{64.78} \\
   
\hline 

  $^{(R)}$\textbf{Intrinsic SCDL} (SVM) & 43.65  \\
  $^{(R)}$\textbf{Extrinsic SCDL} (SVM) & \textbf{64.62}  \\
\hline
  \end{tabular}
   \end{footnotesize}
\end{table}

\noindent We point out the recognition accuracy of $64.62\%$ achieved by our method outperforming state-of-the-art approaches, with the exception of~\cite{huang2016spontaneous}. This shows the effectiveness of the adopted extrinsic SCDL in detecting subtle deformations from 2D landmarks, without any apperance-based information as other approaches in the literature. 

\noindent Compared to the intrinsic approach, it is clear from Table~\ref{tab:casme} that the extrinsic SCDL method is better in recognizing micro-expressions. Recall that the use of extrinsic SCDL to tackle the problem of micro-expression recognition was driven by its good performance in recognizing the contempt emotion, in the CK+ dataset which is characterized by subtle changes along the expression. The obtained results on CASME II hence supports our previous claims. 


\subsection{3D Human Action Recognition}
We evaluate the proposed 3D skeletal representation based on intrinsic SCDL using four benchmark datasets presenting different challenges: Florence3D-Action \cite{SeidenariVBBP13}, UTKinect-Action \cite{xia2012view}, MSR-Action 3D \cite{Li10}, and the large scale NTU-RGB+D \cite{Shahroudy_2016_CVPR}. The obtained recognition accuracies are discussed in section~\ref{res} with respect to Riemannian approaches and other recent approaches that used 3D skeletal data. In addition, we will compare it to the extrinsic SCDL approach presented in section~\ref{sec:4} that we adapted here to the 3D case.  

\begin{itemize}

\item \textbf{Florence3D-Action} dataset \cite{SeidenariVBBP13} consists of 9 actions performed by 10 subjects. Each subject performed every action two or three times for a total of 215 action sequences. The 3D locations of 15 joints collected using the Kinect sensor are provided. The challenges of this dataset consist of the similarity between some actions and also the high intra-class variations as same action can be performed using left or right hand. 

\item \textbf{UTKinect-Action} dataset \cite{xia2012view} consists of 10 actions performed twice by 10 different subjects for a total of 199 action sequences. The 3D locations of 20 different joints captured with a stationary Kinect sensor are provided. The main challenge of this dataset is the variations in the view point. 

\item \textbf{MSR-Action 3D} dataset \cite{Li10} consists of 20 actions performed by 10 different subjects. Each subject performed every action two or three times for a total of 557 sequences. The 3D locations of 20 different joints captured with a depth sensor similar to Kinect are provided with the dataset. This is a challenging dataset because of the high similarity between many actions (\emph{e.g.}, {\it hammer} and {\it hand catch}).

\item \textbf{NTU-RGB+D \cite{Shahroudy_2016_CVPR}} is one of the largest 3D human action recognition datasets. It consists of $56,000$ action  clips of $60$ classes. $40$ participants have been asked to perform these actions in a constrained lab environment, with three camera views recorded simultaneously. Each Kinect sensor estimates and records $25$ joints coordinates reported in the 3D camera's coordinate system.
\end{itemize}

\subsubsection{Experimental Settings}
For the first three datasets, we followed the cross-subject test setting of \cite{Wang12}, in which half of the subjects was used for training and the remaining half was used for testing. Reported results were averaged over ten different combinations of training and test data. For Florence3D-Action and UTKinect-Action datasets, we followed an additional setting for each: Leave-one-actor-out (LOAO) \cite{SeidenariVBBP13,wang2016mining} and Leave-one-sequence-out (LOSO) \cite{xia2012view}, respectively. For MSR-Action3D dataset, we also followed \cite{Li10} and divided the dataset into three subsets AS1, AS2, and AS3, each consisting of 8 actions, and performed recognition on each subset separately, following the cross-subject test setting of \cite{Wang12}. The subsets AS1 and AS2 were intended to group actions with similar movements, while AS3 was intended to group complex actions together. In all experiments, we performed recognition based on the two classification schemes presented in section~\ref{sec:5}.

For NTU-RGB+D, the authors of this dataset recommended two experimental settings that we follow: 1) Cross-subject (X-Sub) benchmark with 39,889 clips from 20 subjects for training and 16,390 from the remaining subjects for testing; 2) Cross-view (X-View) benchmark with 37,462 and 18,817 clips for training and testing. Training clips in this setting come from the camera views 2 and 3 while the testing clips are all from the camera view 1. For this dataset, we construct dictionaries using the kernel clustering approach presented in section~\ref{sec:clustering}. Regarding sparse coding of actions that present interactions, we perform sparse coding of each skeleton separately. Further, we compute displacement vectors, as described in section~\ref{sec:micro} for micro-expressions, and fuse them with SCDL features. Finally, we perform temporal modeling and classification using Bi-LSTM.

\subsubsection{Results and discussions}
\label{res}
 \begin{table*}[!htb]
   \centering
    \caption{Overall recognition accuracy (\%) on MSR-Action 3D, Florence Action 3D, and UTKinect 3D datasets. In the first column: $^{(R)}$:~Riemannian approaches; $^{(N)}$:~other recent approaches; Last row:~our approach.  }
   \label{tab:ComRiem}
 \begin{tabular}{|{c}||{c}|{c}||{c}|{c}||{c}|{c}|}
 \hline
 \textbf{Dataset} & \multicolumn{2}{c||}{\textbf{MSR-Action 3D}} & \multicolumn{2}{c||}{\textbf{Florence 3D}} & \multicolumn{2}{c|}{\textbf{UTKinect 3D}} \\
  \hline
 \textbf{Protocol} & \textbf{Half-Half} & \textbf{3 Subsets} & \textbf{Half-Half} & \textbf{LOAO} & \textbf{Half-Half} & \textbf{LOSO}  \\
 \hline

 $^{(R)}$ T-SRVF on Lie group \cite{anirudh2015elastic} & 85.16 & -- & 89.67 & -- & 94.87 & --
 \\
 $^{(R)}$ T-SRVF on $\cal S$ \cite{BenAmor:2016} & 89.9 & -- & -- & -- & -- & -- 
 \\
 $^{(R)}$ Lie Group \cite{Chellappa-CVPR-2014} & 89.48 & 92.46 & 90.8 & -- & 97.08 & -- 
 \\
 $^{(R)}$ Rolling rotations \cite{vemulapalli2016rolling} & -- & -- & 91.4 & -- & -- & --
 \\
 $^{(R)}$ Gram matrix \cite{kacem2018novel} & -- & -- & -- & 88.85 & -- & \textbf{98.49}
 \\
 \hline 
 $^{(N)}$ Graph-based \cite{wang2016graph}   & -- & -- & -- & 91.63 & 97.44 & --
 \\
 $^{(N)}$ ST-LSTM \cite{liu2016spatio}  & -- & -- & -- & -- & 95.0 & 97.0
 \\
 $^{(N)}$ JLd+RNN \cite{7926607} & -- & -- & -- & -- & 95.96 & --
 \\
 $^{(N)}$ SCK+DCK \cite{koniusz2016tensor} & \textbf{91.45} & 93.96 & \textbf{95.23} & -- & \textbf{98.2} & -- 
 \\
 $^{(N)}$ Transition-Forest \cite{garciatransition} & -- & \textbf{94.57} & -- & 94.16 & -- & --   \\
 \hline
  $^{(R)}$ \textbf{Extrinsic SCDL} (SVM)  & 82.52 & 88.53 & 85.76 & 89.03 & 93.97 & 94.97
  \\
 \hline 
 
  $^{(R)}$ \textbf{Intrinsic SCDL} (SVM) & \textbf{90.01} & \textbf{94.19} & 92.85 & 92.27 & \textbf{97.39} & 97.50
 \\
  $^{(R)}$ \textbf{Intrinsic SCDL} (Bi-LSTM) & 86.18 & 86.18 & \textbf{93.04} & \textbf{94.48} & 96.89 & \textbf{98.49}
\\
 \hline
   \end{tabular}
 \end{table*}

\newcommand\itemss{9}   
\arrayrulecolor{white} 

\begin{table}[!htb]
  \centering
  \caption{Confusion matrix on the Florence Action 3D dataset.}
  \label{tab:conf_Florence}

\begin{tabular}{cc*{\itemss}{|E}|}
\multicolumn{1}{c}{} &\multicolumn{1}{c}{} &\multicolumn{\items}{c}{\textbf{Predicted}} \\ 
\multicolumn{1}{c}{} & 
\multicolumn{1}{c}{} & 
\multicolumn{1}{c}{\rot{Wave}} & 
\multicolumn{1}{c}{\rot{Drink}} & 
\multicolumn{1}{c}{\rot{Phone}} &
\multicolumn{1}{c}{\rot{Clap}} & 
\multicolumn{1}{c}{\rot{Tight lace}} & 
\multicolumn{1}{c}{\rot{Sit-down}} &
\multicolumn{1}{c}{\rot{Stand-up}} & 
\multicolumn{1}{c}{\rot{Read watch}} & 
\multicolumn{1}{c}{\rot{Bow}} \\ \hhline{~*\itemss{|-}|}
\multirow{\itemss}{*}{\rotatebox{90}{\textbf{Actual}}} 
& A1 & 100  & 0 & 0 & 0 & 0 & 0 & 0 & 0& 0\\ \hhline{~*\itemss{|-}|}
& A2 & 2 & 90 & 7 & 0 & 0 & 0 & 0& 0 & 0\\ \hhline{~*\itemss{|-}|}
& A3 & 0  & 12 & 80 & 0 & 0 & 0 & 0 & 8 & 0\\  \hhline{~*\itemss{|-}|}
& A4 & 0  & 0 & 0 & 87 & 0 & 0 & 0 & 13 & 0\\ \hhline{~*\itemss{|-}|}
& A5 & 0  & 0 & 0 & 0 & 93 & 0 & 0 & 3 & 3\\  \hhline{~*\itemss{|-}|}
& A6 & 0  & 0 & 0 & 0 & 0 & 99 & 0 & 1 & 0\\ \hhline{~*\itemss{|-}|}
& A7 & 0  & 0 & 0 & 0 & 0 & 0 & 100 & 0& 0\\ \hhline{~*\itemss{|-}|}
& A8 & 1  & 7 & 5 & 2 & 0 & 0 & 0 & 85 & 0\\ \hhline{~*\itemss{|-}|}
& A9 & 0  & 0 & 0 & 0 & 0 & 0 & 0 & 0 & 100 \\ \hhline{~*\itemss{|-}|}
\end{tabular}
\end{table}

\arrayrulecolor{black} 

\noindent \textbf{Comparison to existing Riemannian representations on MSR-Action, Florence3D and UTKinect datasets} -- The first row of methods in Table~\ref{tab:ComRiem} reports the recognition results of different Riemannian approaches. Since in \cite{BenAmor:2016} human actions are also represented as trajectories in the Kendall's shape space, we report additional results of \cite{BenAmor:2016} on Florence3D and UTKinect datasets to give more insights about the strength of our coding approach compared to the method of \cite{BenAmor:2016}. In Table~\ref{tab:ComRiem}, it can be seen that we obtain better results than all Riemannian approaches on the three datasets. We recall that one common drawback of these methods is to map trajectories on manifolds to a reference tangent space, where they compute distances between different points (other than the tangent point). This may introduce distortions, especially when points are not close to the reference point. However, our method avoids such a non-trivial problem as coding of each shape is performed on its attached tangent space and the only measures that we compute are with respect to the tangent point.
Now, we discuss our results obtained with the first classification scheme, \emph{i.e.}, DTW-FTP-SVM, similarly used in \cite{anirudh2015elastic,Chellappa-CVPR-2014,vemulapalli2016rolling}. In the three datasets, it is clearly seen that our approach outperforms existing approaches when using the same classification pipeline, which shows the effectiveness of our skeletal representation. For instance, we highlight an improvement of  $1.73\%$ on MSR-Action 3D (following protocol \cite{Li10}) and $1.45\%$ on Florence3D-Action.

\noindent Now, we discuss the results we obtained using Bi-LSTM. Note that although we do not perform any preprocessing on the sequences of codes before applying Bi-LSTM, our approach still outperforms existing approaches on Florence3D, with $1.64\%$ higher accuracy. However, it performs less well on UTKinect yielding an average accuracy of $96.89\%$ against $97.08\%$ obtained in \cite{Chellappa-CVPR-2014}. In MSR-Action 3D, our approach performs better than the method of \cite{anirudh2015elastic} using the first protocol. Note that in \cite{anirudh2015elastic}, results were averaged over all 242 possible combinations. However, our average accuracy is lower than other approaches following both protocols on this dataset (around $3.5\%$ in the first and $0.62\%$ in the second). Here, it is important to mention that data provided in MSR-Action 3D are noisy \cite{LOPRESTI2016130}. As a consequence, using Bi-LSTM without any additional processing step to handle the noise (\emph{e.g.}, FTP) could not achieve state-of-the-art results on this dataset.

\vspace{0.2cm}

\noindent \textbf{Comparison to State-of-the-art} -- We discuss our results with respect to recent non Riemannian approaches. In all datasets, our approach achieved competitive results.

\vspace{0.2cm}

\noindent \textbf{Florence3D-Action} -- On this dataset, our method outperforms other methods using Bi-LSTM in the case of LOAO protocol, as shown in Table~\ref{tab:ComRiem}. However, using the second protocol, it is $2.19\%$ lower than \cite{koniusz2016tensor}. The authors of \cite{koniusz2016tensor} combine two kernel representations: sequence compatibility kernel (SCK) and dynamics compatibility kernel (DCK) which separately achieved $92.98\%$ and $92.77\%$, respectively. The proposed approach achieves good performance for most of the actions. However, the main confusions concern very similar actions, \emph{e.g.}, \textit{Drink from a bottle} and \textit{answer phone}, as shown in the confusion matrix in Table \ref{tab:conf_Florence}.

\noindent \textbf{UTKinect} -- Following the LOSO setting, our approach achieves the best recognition rate, yielding an improvement of $2.49\%$ compared to the method of \cite{liu2016spatio}, which is based on an extended version of LSTM. For the second protocol, our best result is competitive to the accuracy of $98.2\%$ obtained in \cite{koniusz2016tensor}. Considering the main challenge of this dataset, \emph{i.e.}, variations in the view point, our approach confirms the importance of the invariance properties gained by adopting the Kendall's representation of shape, hence the relevance of the resulting functions of codes generated using the geometry of the manifold.

\noindent \textbf{MSR-Action 3D} --
For the experimental setting of \cite{Li10}, our best result is competitive to recent approaches. In particular, on AS3, we report the highest accuracy of $100\%$. This result shows the efficiency of our approach in recognizing complex actions, as AS3 was intended to group complex actions together. On AS1, we achieved one of the highest accuracies ($95.87\%$). However, our result on AS2 is about $8.9\%$ lower than state-of-the-art best result. This shows that our approach performs less well when recognizing similar actions, as AS2 was intended to group similar actions together. Although our best result is slightly higher than \cite{koniusz2016tensor}, it is lower than the same method when following the experimental setting of \cite{Wang12a}. This shows that our approach performs better in recognition problems with less classes.

 \begin{table}[!htb]
   \centering
    \caption{Overall recognition accuracy (\%) on NTU-RGB+D following the X-sub and X-view protocols. In the first column: $^{(R)}$:~Riemannian approaches; $^{(RN)}$:~RNN-based approaches; $^{(CN)}$:~CNN-based approaches.}
   \label{tab:NTU}
 \begin{tabular}{|{c}|{c}|{c}|}
   \hline
 \textbf{Protocol} & \textbf{X-sub} & \textbf{X-view} \\
 \hline
$^{(R)}$ Lie Group \cite{veeriah2015differential} & 50.1 & 52.8 \\
HB-RNN-L \cite{du2015hierarchical} & 59.1 & 64.0 \\
$^{(R)}$ Deep learing on $SO(3)^n$ \cite{huang2017deep}  & 61.3 & 66.9 \\
$^{(RN)}$Deep LSTM \cite{shahroudy2016ntu} & 60.7 & 67.3 \\
$^{(RN)}$Part aware-LSTM \cite{shahroudy2016ntu} & 62.9 & 70.3 \\
$^{(RN)}$ST-LSTM+Trust Gate \cite{liu2016spatio} & 69.2 & 77.7 \\
$^{(RN)}$View Adaptive LSTM \cite{zhang2017view} & 79.4 & 87.6 \\
\hline
$^{(CN)}$Temporal Conv \cite{kim2017interpretable} & 74.3 & 83.1 \\
$^{(CN)}$C-CNN+MTLN \cite{ke2017new} & 79.6 & 84.8 \\
$^{(CN)}$ST-GCN \cite{yan2018spatial} & \textbf{81.5} & \textbf{88.3} \\
 \hline 
  $^{(R)}$ \textbf{Intrinsic SCDL}  &  \textbf{73.89} & \textbf{82.95}  
\\
 \hline
   \end{tabular}
 \end{table}
 
\begin{figure}[!ht]
  \centering
   \includegraphics[width=.98\linewidth]{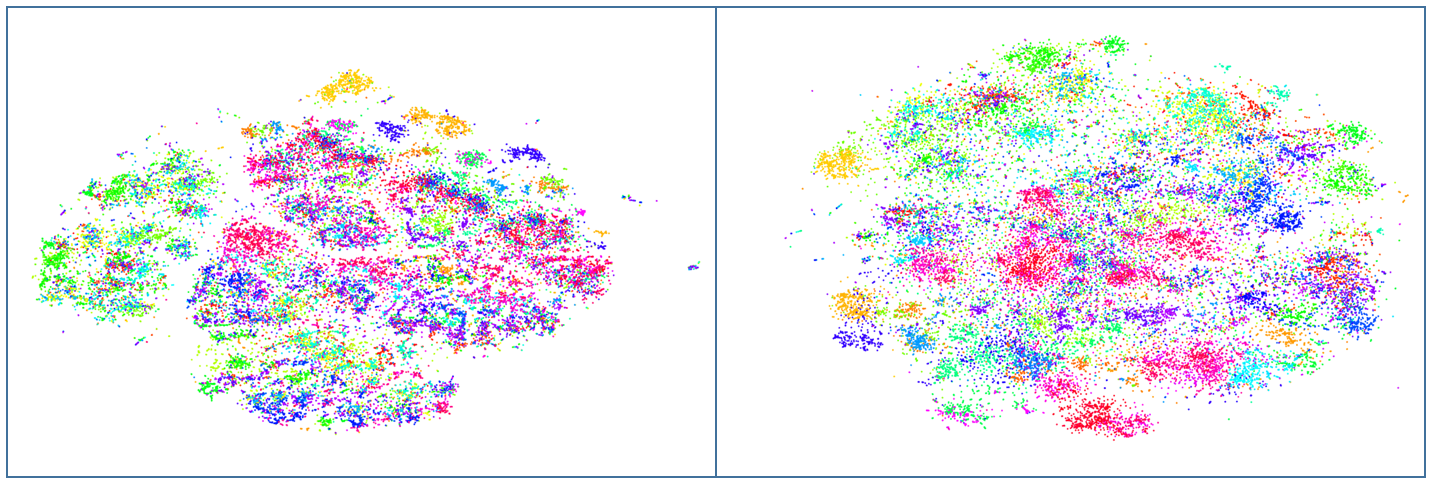}
  \caption{Visualization of 2-dimensional features of the NTU-RGB+D dataset. Left: original data. Right: the corresponding SCDL features. Each class is represented by a different color. This figure is better seen in colors.}
   \label{Fig:t_sne}
\end{figure}

\noindent\textbf{NTU-RGB+D} -- We report the obtained results for this dataset in Table.~\ref{tab:NTU}. For both benchmarks, X-view and X-sub, our approach remarkably outperforms other Riemannian representations. For instance, it outperforms the Lie group representation by 23\% and 30\% on X-sub and X-view protocols. It also surpasses the deep learning on Lie groups method by 12\% and 16\%. This could demonstrate the ability of our approach to deal with large scale datasets compared to conventional Riemannian approaches. Besides, our method outperforms RNN-based models, HB-RNN-L, Deep LSTM, PA LSTM and ST-LSTM+TG, with the exception of~\cite{zhang2017view}. Knowing that we also used an RNN-based model (Bi-LSTM) for temporal modeling and classification, this shows the efficiency of our action modeling. In fact, sparse features obtained after SCDL in Kendall's shape space are remarkably more discriminative than the original data. In order to have a better insight into their corresponding data distributions, we used the t-distributed stochastic neighbor embedding (t-SNE)\footnote{t-SNE is a nonlinear dimensionality reduction technique that allows for embedding high-dimensional data into two or three dimensional space, which can then be visualized in a scatter plot.} to visualize original data and SCDL features. From Fig.~\ref{Fig:t_sne}, we can observe that the SCDL features are better clustered than the original data in terms of class labels (colors in the figure). Besides, it is worth noting that SCDL is an efficient denoising tool, which is an important advantage when dealing with the often-noisy skeletons extracted with the Kinect sensor.   

\vspace{0.2cm}

\vspace{0.2cm}

\noindent \textbf{Comparison to extrinsic SCDL} -- To further evaluate the strength of the proposed intrinsic approach in the context of 3D action recognition, we compare it to the extrinsic SCDL method presented in section~\ref{sec:4} that we adapted here to the 3D Kendall's space. As explained in section~\ref{sec:3}, for 2D shapes, the authors in~\cite{jayasumana2013framework} proved the positive definiteness of the Procrustes Gaussian kernel which is based on the full Procrustes distance. For 3D shapes, we adapted the extrinsic SCDL formulation (presented in section~\ref{sec:4}) by applying the Procrustes Gaussian kernel, in which we also adapted the full Procrustes distance to 3D shapes as $d_{FP}(\bar{Z}_1,\bar{Z}_2)=sin(\theta)$ (see section 4.2.1 of \cite{dryden2016statistical}) ($\theta$ is the geodesic distance defined in section~\ref{KendallTheory}). Experimentally, we checked the positive definiteness of the adapted kernel and found out that it is only positive definite for some values of $\sigma$. We empirically chose $0.1$ for Florence3D, $0.2$ for UTKinect, and $0.5$ for MSR-Action 3D, as to have valid positive definite kernels. Results reported in Table~\ref{tab:ComRiem} show superiority of the proposed intrinsic method. Note that the accuracy obtained for UTKinect was updated compared to \cite{tanfous2018coding} by testing further values of $\sigma$.

\subsection{Ablation study}
We examine the effectiveness of the proposed Kendall SCDL schemes by performing different baseline experiments on the action datasets. In addition, we evaluate the performance of our approach in the task of facial expression recognition when using two different facial landmark detectors.

\vspace{0.2cm}

\noindent\textbf{A. Kendall's shape representation} -- We evaluate the necessity of the Kendall's shape projection. To this end, we perform temporal modeling and classification on raw data, after a scale and translation normalization, against their application on Kendall SCDL features. On NTU-RGB+D, we applied Bi-LSTM while on Florence 3D, MSR Action 3D and UTKinect, we applied the pipeline DTW-FTP-SVM. Performances are reported in the second and fourth rows of Table \ref{tab:Ablation}. In all datasets, improvements are remarkably gained with the Kendall's space projection. This is clearly seen in particular on the large scale NTU-RBD+D dataset which presents different view-variations and where the improvement is more than 26\%.

 \begin{table}[!htb]
   \centering
    \caption{Evaluation of the Kendall's shape space representation.}
    \begin{tabular}{|c|c|c|c|c|}
\hline

    \textbf{Approach} & NTU-RGB+D &  Florence &  MSR 3D &  UTKinect \\
    \hline
     Raw data  & 56.5  & 84.29  & 87.36  & 92.67  \\
    \hline
     Linear SCDL & 79.20 & 87.94 & 89.23 & 93.58  \\
     \hline
     \textbf{Kendall SCDL} & \textbf{82.95} & \textbf{92.85} & \textbf{90.01} &  \textbf{97.5} \\
      \hline
       
\end{tabular}
\label{tab:Ablation}
 \end{table}

\vspace{0.2cm}

\noindent\textbf{B. Nonlinear SCDL} -- In this experiment, we evaluate the importance of the nonlinear formulation of SCDL that we applied on Kendall's space. For that, we compare it to the use of linear SCDL, i.e., by solving for Eq.\ref{eq:SC-linear}. Obtained results on the four action recognition datasets, reported in the third row of Table~\ref{tab:Ablation}, clearly show the interest of accounting for the nonlinearity of the manifold when applying SCDL.

\begin{table}[!htb]
   \centering
    \caption{Classification performances when using different landmark detectors.}
\begin{tabular}{|c|c|c|}
\hline
    \textbf{Landmark detector} & Oulu-CASIA dataset &  CK+ dataset \\
    \hline
     Chehra \cite{6909636} - \textbf{49 landmarks} & \textbf{76.41}  & \textbf{93.68}  \\
     \hline
     Openface \cite{baltrusaitis2018openface} - \textbf{49 landmarks}  & 70.85 & 83.73 \\
      \hline
     Openface \cite{baltrusaitis2018openface} - 68 landmarks & 71.26 & 82.92\\
       \hline
       
\end{tabular}
\label{tab:AblationFER}
\end{table}

\vspace{0.2cm}

\noindent\textbf{C. Facial landmark detectors} -- The task of facial expression recognition from landmark data relies essentially on the accuracy of the landmark detector. In this experiment, we evaluate the performance of the landmark detector that we used in our experiments (\textit{i.e.}, Chehra \cite{6909636}) by comparing it to the newly-released Openface2.0 \cite{baltrusaitis2018openface}, which gives the option of extracting either 49 or 68 landmarks.  In Table~\ref{tab:AblationFER}, we report the classification accuracy obtained by applying the pipeline DTW+FTP+SVM on raw landmark data (after a simple scale and translation normalization). Results obtained on CK+ and Oulu-CASIA datasets clearly show a better performance using landmarks extracted with the Chehra detector.

\subsection{Discussion}
The Kendall's shape representation has proven the efficiency of adopting a view-invariant analysis of the given data. Because of the nonlinearity of the Kendall's manifold, intrinsic and extrinsic solutions of SCDL were comprehensively studied and compared. Regarding the extrinsic solution, the advantage of embedding data from the Kendall's shape space to RKHS is twofold. First, the latter is vector space, thus it enables the extension of the well established linear SCDL on the nonlinear Kendall's space. Second, embedding a lower dimensional space in a higher dimensional one gives a richer representation of the data and helps extracting complex patterns. However, to define a valid RKHS, the kernel function must be positive definite according to Mercer's theorem.
\noindent On one hand, for the 2D Kendall's space, we have used the Procrustes Gaussian Kernel which is positive definite and shown that for the task of 2D macro facial expression recognition, extrinsic SCDL performs better than intrinsic SCDL. We argue that this is due to the kernel embedding. For instance, we highlight the clear improvement in recognizing the \emph{contempt} emotion in the CK+ dataset. The latter is characterized with subtle deformations that are well captured using the extrinsic approach. This has drove us to evaluate it on the task of 2D micro-expression recognition where the shape deformation along expressions are know to be subtle as well. As expected, the performance of extrinsic SCDL was promising.
\noindent On the other hand, for the 3D Kendall's space, a positive definite kernel function has not been proposed in the literature. Nevertheless, adapting the PGk to 3D shapes prevented us from exploring the whole space of $\sigma$ as in this case, this kernel is positive definite for only certain value of this parameter. As a consequence, the performance of extrinsic SCDL in the 3D Kendall's space can be hindered since the quality of the produced codes depends on the value of $\sigma$. We argue that this is the main reason behind the better performance obtained using intrinsic SCDL for the task of 3D action recognition. Besides, intrinsic sparse coding of a shape is performed on its attached tangent space, by mapping atoms into it. Compared to Riemannian approaches of the literature, this avoids the common drawback of mapping points to a common tangent space at a reference point which may introduce distortions.

\section{Conclusion}
\label{sec:Concl} 
In this paper, we proposed novel representations of human actions and facial expressions based on Riemannian SCDL in Kendall's shape spaces. We represented sequences of landmark configurations as trajectories in the Kendall's space to seek for a view-invariant analysis. To deal with the nonlinearity of this manifold, a Riemannian dictionary is learned from the data and used to efficiently code static shapes. This yield discriminative sparse time-series that are processed and classified in vector space. Intrinsic and extrinsic solutions of SCDL were explored. We showed that while intrinsic SCDL is more suitable in coding 3D skeletal trajectories, extrinsic SCDL is more effective in coding 2D facial expressions. Extensive experiments were conducted for the tasks of 2D macro- and micro- facial expression recognition and 3D action recognition. The obtained results are competitive to state-of-the-art solutions showing the relevance of our proposed representations for the given tasks. Furthermore, a comprehensive study comparing the two proposed SCDL solutions was presented in this paper. 

For future works, given the dictionary of shapes, SCDL features can be efficiently reconstructed to obtain a good approximation of original data. This property could be very useful in different tasks where a mapping to the original data space is needed. Examples are the prediction of human motion or the generation of novel actions or facial expressions. In these tasks, one could train a generative or a predictive model with SCDL features. Then the generated (respectively predicted) features will be reconstructed to give rise to data in the original space of human skeletons or landmark faces.


\vspace{-1cm}

\begin{IEEEbiography}[{\includegraphics[width=1in,height=1.25in,clip,keepaspectratio]{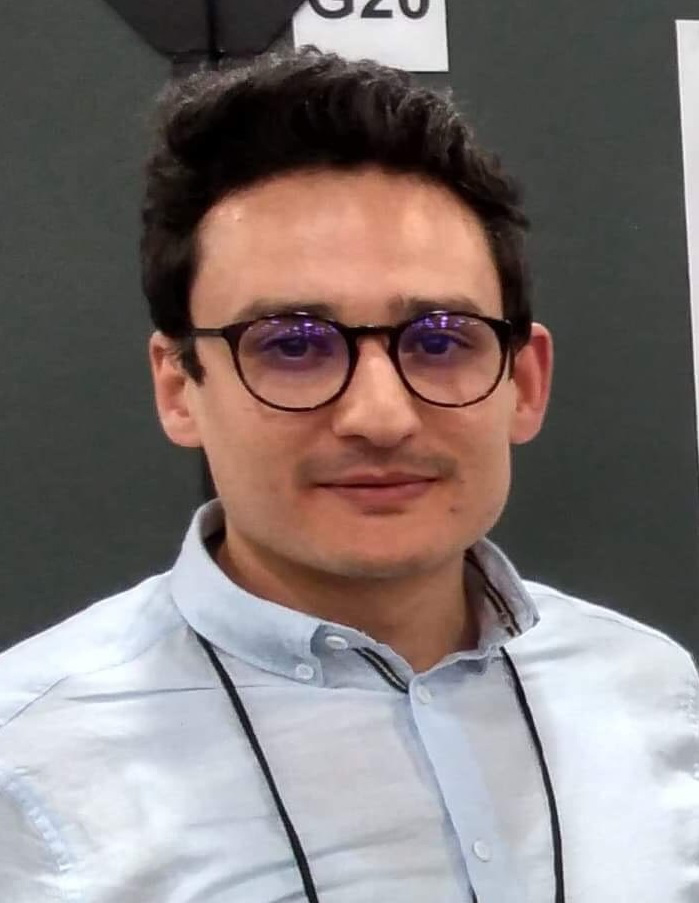}}]{Amor Ben Tanfous} received the engineering degree from the National Institute of Applied Sciences and Technology (INSAT), in Tunisia, in 2016 and the MS double-degree from the National Engineering School of Tunis and the Paris Descartes University, in 2016. He is working toward the PhD degree at Mines-Telecom Institute (IMT) Lille Douai/University of Lille and member of the CRIStAL Lab (CNRS UMR 9189). He is currently interested in solving computer vision problems in the interface of machine learning and shape analysis.
 
\end{IEEEbiography}

\vspace{-1cm}

\begin{IEEEbiography}[{\includegraphics[width=1in,height=1.25in,clip,keepaspectratio]{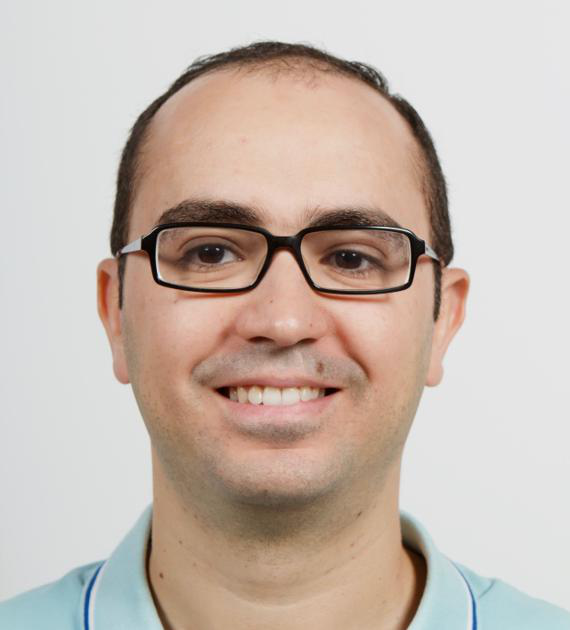}}]{Hassen Drira} received the PhD degree in computer science from the University of Lille1. He is an associate professor of computer science with the Mines-Telecom Institute, IMT Lille Douai and member of the UMR CNRS 9189 Research center CRIStAL, in France. His research interests include pattern recognition, shape analysis and computer vision. He has published several refereed journals and conference articles in these areas.
\end{IEEEbiography}

\vspace{-1cm}

\begin{IEEEbiography}
[{\includegraphics[width=1in,height=1.25in,clip,keepaspectratio]{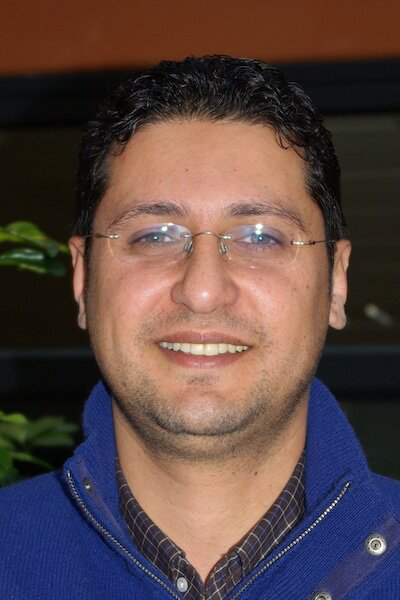}}]{Boulbaba Ben Amor} received the habilitation to supervise reserach (H.D.R.) from the University of Lille, France and the PhD degree in computer science from the Ecole Centrale de Lyon, in 2006. He joined the Inception Institute of Artificial Intelligence (IIAI) in U.A.E as senior scientist, from his full professor position with the Mines-Telecom Institute (IMT) Lille Douai, in France. He is recipient of the prestigious Research Fulbright scholarship (2016-2017). His current research topics lie to 3D shape and their dynamics analysis for advanced human behavior analysis. He is senior member of the IEEE, since 2015.
\end{IEEEbiography}

\end{document}